\documentclass[runningheads]{llncs}
\usepackage[T1]{fontenc}
\usepackage{graphicx}
\usepackage[inkscapelatex=false]{svg}
\usepackage{graphicx}
\usepackage{xcolor}
\usepackage{comment}
\usepackage{amsmath,cite,url}
\usepackage{booktabs}

\newcommand{\etal}{\textit{et al.}}

\newcommand{\tit}[1]{\smallbreak\noindent\textbf{#1.}}

\begin{document}
\title{GAN-based Content-Conditioned Generation
of Handwritten Musical Symbols}
\titlerunning{GAN-based Optical Music Generation}

\author{Gerard Asbert\inst{1}\orcidID{0009-0005-3316-5463} \and
Pau Torras\inst{1,2}\orcidID{0000-0003-0327-9046} \and
Lei Kang\inst{1,2}\orcidID{0000-0002-1962-3916} \and
Alicia Fornés\inst{1,2}\orcidID{0000-0002-9692-5336}\and
Josep Lladós\inst{1,2}\orcidID{0000-0002-4533-4739}}

\authorrunning{G. Asbert et al.}

\institute{Computer Vision Center, Barcelona, Spain 
\email{\{gasbert, ptorras, lkang\}@cvc.uab.cat} \email{\{afornes, josep\}@cvc.uab.es} \and
Department of Computer Science, Universitat Autònoma de Barcelona, Spain\\
}

\maketitle              % typeset the header of the contribution
\begin{abstract}
The field of Optical Music Recognition (OMR) is currently hindered by the scarcity of real annotated data, particularly when dealing with handwritten historical musical scores. In similar fields, such as Handwritten Text Recognition, it was proven that synthetic examples produced with image generation techniques could help to train better-performing recognition architectures. This study explores the generation of realistic, handwritten-looking scores by implementing a music symbol-level Generative Adversarial Network (GAN) and assembling its output into a full score using the Smashcima engraving software. We have systematically evaluated the visual fidelity of these generated samples, concluding that the generated symbols exhibit a high degree of realism, marking significant progress in synthetic score generation.

\keywords{Generative Adversarial Networks  \and Handwritten Musical Symbols \and Content Conditioning \and Musical Score Generation}
\end{abstract}
\section{Introduction}
Optical Music Recognition (OMR) is formally defined as the task of computationally reading music notation in documents \cite{calvo2020understanding}, whether printed or handwritten. From a research point of view, one of the main goals of OMR is constructing faithful computer-processable representations of these music scores, which would enable their study by scholars and their conservation by archivists. Having computers perform this task is interesting because manual transcription of musical documents is a painstaking, costly process that requires many hours, even for relatively simple scores and seasoned musicologists.

The complexity of this task highly increases when addressing historical handwritten scores, particularly those written in CWMN (Common Western Music Notation) from the 18th century onward. Thousands of these musical scores, are scattered throughout the world, only a small percentage of which has been transcribed. This scarcity of data added to the presence of artifacts due to the aging of the paper plummet the performance of OMR models. Multiple authors \cite{calvo2020understanding, torras2024unified} have stated that the scarcity of annotated handwritten musical data is one of the biggest bottlenecks to the improvement of the field. Many have resorted to using printed musical data as a workaround, which can provide an improvement in model performance \cite{baro_handwritten_2020}. However, it remains inferior to the value offered by real handwritten data. Therefore, the generation of synthetic handwritten data seems a viable option to explore. 

For this reason, the aim of this work is to design an effective and robust generative architecture that can generate synthetic handwritten scores to increase the training set for OMR systems, and thus, improve their overall performance. Concretely, we propose to generate isolated symbols using Generative Adversarial Networks (GAN) \cite{goodfellowGenerativeAdversarialNetworks2020}, and use the Smashcima engraving software \cite{mayer2021synthesizing} to create full music sheets. From the extensive evaluation, we observe that the resulting music scores are realistic, paving the way to the use of generated music data for training OMR systems.

%\TODO{Paragraf parlant de l'estructura del document}
As for the structure followed in this work, after outlining the research context and motivations, we review prior work in OMR, handwriting generation and music symbol synthesis. The methodology section presents our adapted Generative Adversarial Network (GAN) architecture, followed by a description of the dataset composition and preprocessing techniques. The experimental setup is then detailed, including hyperparameter selection and training strategies. Finally, evaluation is conducted through qualitative visual assessments and quantitative metrics such as Fréchet Inception Distance, Kernel Inception Distance, and Handwriting Distance.

\section{Previous Work}
Early OMR systems relied on classical computer vision techniques to extract simple symbol primitives from images \cite{rebelo2012optical,tardon2009optical,fornes2007old}. While these methods gave acceptable results for closed domains, the use of deep learning models significantly improved recognition accuracy and generalization on many sub-tasks \cite{calvo2020understanding}. Unfortunately, implementing large-scale learning algorithms is complicated in the field of OMR because of the requirement of large amounts of annotated data, which are not quite there yet \cite{calvo2020understanding,torras2024unified}. Consequently, it has become necessary to explore alternative strategies to obtain samples to train models on, including data augmentation, synthetic dataset generation, and transfer learning from printed data.

%\tit{Handwriting Generation}
%When the generation of synthetic data was not as developed as today, researchers could only rely on basic data augmentation, such as rotations and mirroring, to expand datasets, as seen in \cite{tardon2009optical}. 

Despite the fact that data augmentation \cite{tardon2009optical} can mitigate overfitting, it is inherently limited to producing variations within the original feature space, being unable to generate entirely new samples. The creation and popularization of generative deep learning models surpassed this limitation, as it enables to generate realistic samples and not as resembling of the input image. 
Among the predominant generative models developed during the past decade, including Generative Adversarial Networks (GANs) \cite{goodfellowGenerativeAdversarialNetworks2020}, Diffusion Models (DFs) \cite{hoDenoisingDiffusionProbabilistic2020}, Variational Autoencoders (VAEs) \cite{kingmaAutoEncodingVariationalBayes2022}, and Transformers \cite{mehmood2023deep, bond2021deep}, GAN architectures have been among the most popular in the field of Handwritten Text Recognition (HTR). Indeed, HTR is a much more developed field than OMR, especially in the generation of synthetic data. Numerous studies \cite{kang2020ganwriting,kangContentStyleAware2022,zdenekJokerGANMemoryEfficientModel2021,zdenekHandwrittenTextGeneration2023,elanwar2024generative} have leveraged Generative Adversarial Networks (GANs) for synthesizing handwritten words, benefiting from the extensive availability of training data—far exceeding the datasets available for handwritten OMR. In this context, the work of Kang \etal \cite{kang2020ganwriting} was among the first to apply a GAN architecture in the field of HTR, yet it managed to generate highly realistic handwritten text, and thus showing a promising research direction for related domains, such as handwritten music symbols. 

%\tit{Handwritten Music Generation}
However, very few of the above mentioned techniques have been explored in the context of Optical Music Recognition. VAEs \cite{kingmaAutoEncodingVariationalBayes2022} and Adversarial Autoencoders (AAEs) \cite{makhzaniAdversarialAutoencoders2016} have been tested in \cite{havelka2023symbol} showing acceptable results. Also in the field of OMR, Shatri et al.\cite{shatri2024synthesising} and Tirupati et al.\cite{tirupati2024crafting} propose the use of GANs for line-level music scores. In both works, the generated images display notable realism, however, there is still room for improvement, especially in the generation of small and detailed symbols, such as accidentals and rests. Finally, it is worth to mention that there are other types of contributions to the field of data generation that do not necessarily involve deep learning models. The Smashcima software \cite{mayer2021synthesizing}, created by Mayer and Pecina, consists of an architecture that takes existing individual music symbols and engraves them into realistic music sheets. However, it does not generate new music symbols, which means that, in terms of handwriting style variability, it is limited to the styles of the existing set of music symbols.

%We will be using this work to test how our synthetic music symbols, when placed into sheets, react to the different metrics selected in this project.
%
%\tit{Generation of Handwritten Text}
%Handwritten Text Recognition (HTR) is a much more developed field than OMR, especially in the generation of synthetic data. Numerous studies \cite{elanwar2024generative} have leveraged Generative Adversarial Networks (GANs) for synthesizing handwritten words, benefiting from the extensive availability of training data—far exceeding the datasets available for handwritten OMR. The work \cite{kang2020ganwriting} by Kang \etal, was among the first to apply a GAN architecture in the field of HTR, yet it managed to generate highly promising results, thus, demonstrating its potential if applied in related domains, such as OMR. 

In this work, and inspired by the above findings, we propose to combine the benefits of GANs and the Smashcima software to generate more realistic and varied music scores. Concretely, we propose a content-conditioned GAN architecture to generate music symbols to be fed into the Smashcima software. Thus, we can generate music scores without limitations regarding contents and handwriting styles while avoiding the need of well-aligned datasets for training -- we can rely on symbol-level annotations from widespread datasets instead, which are easier to obtain, and generate smaller and more controlled classes, which is computationally more efficient.

\section{Methodology}

\subsection{Preliminaries}
The architecture of a basic GAN \cite{goodfellowGenerativeAdversarialNetworks2020} comprises two main networks: The Generator $G(\vec{z};\vec{\theta_g})$ and the Discriminator $D(\vec{x}; \vec{\theta_d})$, which are differentiable functions parameterized by some learnable weights $\vec{\theta_{[\cdot]}}$. The Generator produces an image conditioned by some noise vector $\vec{z}$, while the Discriminator computes the probability that an image $\vec{x}$ is a real sample or is produced by the Generator. These two networks' parameters are optimized jointly by minimizing the discriminator loss
\begin{equation}
    \mathcal{L}_d = \log(1-D(G(\vec{z}))).
\end{equation}
Essentially, training follows an adversarial process: the Discriminator minimizes its classification error, while the Generator maximizes it, aiming to generate images indistinguishable from real ones. 

This min-max optimization has its benefits, as it fosters continuous improvement in both networks while not relying on a pixel-level loss function. Nevertheless, it also has its drawbacks, the most prominent being, that balancing the simultaneous learning of both architectures can be difficult -- if one network significantly outperforms the other, the adversarial dynamic collapses. This scenario may occur if the Discriminator quickly reaches near-perfect accuracy, which can lead to vanishing gradients for the Generator. Alternatively, if the Generator produces non-realistic samples while the Discriminator has not learned to distinguish them correctly, the Discriminator’s loss quickly saturates and stops providing useful feedback. 

\subsection{Baseline Model}

We base our work on the GANWriting model by Kang \etal  \cite{kang2020ganwriting}. This model expands on the basic premise of a GAN, adapting it to the idiosyncrasy and requirements of handwriting generation. The primary distinction from a conventional GAN is the ability to condition the output to adapt to a specific writing style. Rather than using a random vector as input, the Generator receives an encoded representation of a set of images from the same author, forcing the GAN to learn their specific calligraphic style. The secondary main distinction is that the model can be conditioned to generate images with a specific textual content. Additionally, additive noise is applied to the feature space representing the calligraphic style before concatenating the two content encoding vectors, forcing the appearance of alterations on the output.

To support this finer-grained generation process, the model employs three distinct loss functions to guide the training process. The first one is the Discriminator Loss from standard GANs. The second one is the Word Recognizer Loss, which is computed by passing the generated image through a word recognition model, encouraging the Generator to produce images that accurately contain the word specified in the input string. Lastly, the Writer Classifier Loss incentivizes the Generator to generate images that exhibit a stylistic resemblance to the handwriting of the specified author from the input set of images.

\begin{comment}
(O.H.E in Fig. ~\ref{fig:arquitectura}). This conditioning mechanism ensures that the generated output corresponds to the specified symbol class, rather than a randomly sampled symbol from the training set. Additionally, to further enforce class consistency, the architecture includes a music symbol classifier (S in Fig. ~\ref{fig:arquitectura}), which computes a classifier loss. This loss penalizes the Generator if the generated symbol does not match the intended class, thereby improving the accuracy of the generated symbols.
\end{comment}

\subsection{Adaptation to Music Symbol Generation} \label{subsec:symbgen}

\begin{figure*}[htbp]
    \centerline{\includegraphics[width=\textwidth]{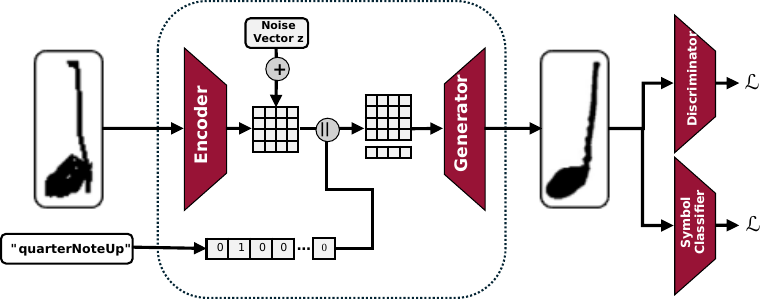}}
    \caption{Architecture of the proposed handwritten musical symbol generation model. The input to the generator is a style example and the symbol class encoded as a one-hot encoded vector. The architecture is trained combining a standard GAN discriminator loss and a symbol classification loss, each produced by two sub-networks.}
    \label{fig:arquitectura}
\end{figure*}

Our architecture, as can be observed in Fig. ~\ref{fig:arquitectura}, replicates the content conditioning in \cite{kang2020ganwriting} with some necessary changes. Given that writer identification data is not available for all of the datasets in use, we restrict the style input to a single image. Furthermore, the content conditioning sequence is replaced by a single one-hot encoded vector, given that individual symbols are generated on every run instead of full words. These changes imply the modification of the loss function of the model w.r.t. the baseline implementation and thus the modification of specific components of the model.

We employ the same Discriminator network proposed in \cite{kang2020ganwriting}. This architecture consists of an initial convolutional layer, followed by six residual blocks with Leaky ReLU activations and Average Pooling. A final binary classification layer is used to distinguish between real and generated images. The standard GAN Discriminator loss $\mathcal{L}_d$ is computed from this module.

For the content loss, the text recognition network is replaced by a CNN-based symbol classifier $C(\vec{x};\vec{\theta_c})$, which consists of three convolutional layers with batch normalization. Each of these blocks is in turn followed by ReLU activation and max pooling to progressively downsample the spatial dimensions while increasing feature depth. The extracted feature maps are flattened and fed into a final classification layer that outputs predictions over our vocabulary of musical symbols. Denoting the final predicted distribution over the output class vocabulary $V$ as $C(\vec{x})=\hat{y}$ and the ground truth distribution as $y$, the final Classification loss $\mathcal{L}_c$ is formally defined as
\begin{equation}
    \mathcal{L}_c=\text{KL}(\hat{y}, y) = \sum_{\forall v \in V} y(v) \cdot \log \frac{y(v)}{\hat{y}(v)}.
\end{equation}

A sweep was performed to test if the above specified Kullback–Leibler divergence loss was the right choice compared to the usual cross-entropy loss, for which the former gave slightly better results.

Given that there is no writer identification ground truth in all of the training datasets, the writer identification loss is not used for this model. The final loss function for the full model is thus
\begin{equation}
    \mathcal{L_{\text{Model}}}=\alpha\mathcal{L}_d+\beta\mathcal{L}_c,
\end{equation}
where both $\alpha$ and $\beta$ are hyperparameters that allow weighting the importance of each of the sub-modules.

\section{Datasets}
The data used for training the GAN consists of images of real handwritten musical symbols, which have been extracted from multiple sources: 
\tit{MUSCIMA++ \cite{hajivc2017muscima++}:} A collection of 140 page-level modern handwritten score samples with pixel-level symbol annotations. It contains examples of most classes. 
\tit{Fornes Dataset \cite{fornes2007old}:} A symbol-level dataset containing 2128 examples of clefs and 1970 examples of accidentals. Originally intended for classification.
\tit{Homus \cite{calvo2014recognition}:} A broad ``online'' (as opposed to rasterized; containing stroke information) symbol database containing 15400 samples authored by 100 different musicians. It contains examples of most classes. 
\tit{Capitan Collection \cite{calvo2016two}:} A collection of symbols extracted by tracing real mensural scores. Some of the classes are compatible with CWMN.

The combination of these datasets is not straightforward, given that their symbol definitions are not directly compatible. A pre-processing step is performed to address inconsistencies in naming conventions for objects pertaining to the same class (e.g. \texttt{Eight-Rest} from Homus or \texttt{Rest8th} from MUSCIMA++ to a generic \texttt{eightrest} class) and to address granularity differences between the classes of the datasets (e.g. converting objects in the \texttt{halfNote} class into \texttt{halfNoteUp} and \texttt{halfNoteDown}).

Due to the scarcity of samples and high imbalance found in the initial class distribution, data augmentation techniques are applied on the training symbol dataset. New images are created by rotating dataset samples by $\pm 10^\circ$ or mirror flipping them in those cases where semantic integrity is preserved, thus re-balancing some of the underrepresented classes. A threshold of 3,000 images per class is set to ensure an adequate amount of training data for reliable model performance, resulting in a final collection of 49 classes.

Additional line-level samples are required in order to evaluate the full generation pipeline -- generation of symbols and engraving. The MUSCIMA++ dataset is used as gold standard, since it is the only one of the sourced datasets that contains full-line images as well as symbol annotations. This ensures that comparison is performed against samples that are reasonably in-distribution.

\section{Experimental Setup}
\subsection{Hyperparameter Tuning} \label{subsec:hyperparam}

The learning rates for the discriminator, generator, and recognizer are set to 1e-5, 1e-4, and 1e-5, respectively, based on optimal performance observed during a hyperparameter sweep. We employ a batch size of 16, chosen to balance computational efficiency with stable gradient updates.

To regulate training dynamics, we apply different weighting factors to the loss components. The discriminator loss is weighted at $\alpha=1.0$, while the recognizer loss is assigned a weight of $\beta=2.5$ to emphasize accurate classification of generated symbols. Additionally, a noise penalty term with a weight of 3.0 is introduced to encourage diversity in the generated samples, mitigating mode collapse. 

\subsection{Additional Modifications} \label{subsec:additional}

\begin{figure}[h]
    \centering
    \begin{minipage}{0.48\columnwidth}
        \centering
        (a) \\  % Label above the image
        \makebox[\linewidth]{\includegraphics[width=1.1\linewidth]{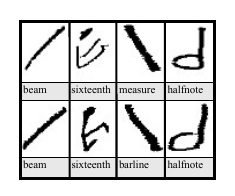}}
    \end{minipage}
    \hfill
    \begin{minipage}{0.48\columnwidth}
        \centering
        (b) \\  % Label above the image
        \makebox[\linewidth]{\includegraphics[width=1.1\linewidth]{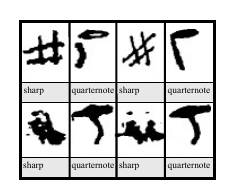}}
    \end{minipage}
    \caption{(a) An example of an average generation. (b) An example of a faulty generation. Top row: Input symbols | Bottom row: Generated symbols.}
    \label{fig:hw1}
\end{figure}

Applying the described GAN architecture to handwritten data produces already acceptable synthetic symbols, as shown in Fig. ~\ref{fig:hw1}.a. Nevertheless, certain generated examples, as illustrated in Fig. ~\ref{fig:hw1}.b, display some issues that require further refinement. 

The primary challenge is the GAN’s difficulty to generate handwritten-looking images for the more complex symbols, concretely the \texttt{accidentalSharp} and \texttt{gClef} classes. As it can be observed in Fig. ~\ref{fig:hw1}.b, results often do not resemble the input symbol. Our current hypothesis suggests that this issue is caused by the extreme variability in the handwritten representations of these symbols, which may hinder the model’s ability to learn a consistent mapping.

To address this issue, two potential solutions are explored. The first solution involves implementing symbol-specific training steps. After 150 standard steps -- that is, where each batch contains randomly selected symbols -- the GAN undergoes 50 steps where the batch consists exclusively of the symbols that prove more challenging to generate, allowing for a more focused training on them.

The second solution involves the symbol classifier and its corresponding loss. The issue arises from the fact that the symbol classifier correctly labels disorted symbols because they resemble the real symbol class more closely than the other classes. Consequently, the loss does not reflect the presence of distortions, preventing the GAN from learning to correct them. Our hypothesis suggests that the classifier needs to be more stringent, as a result, several pictures of the distorted symbols were assigned to new ``bad'' classes (in this case, \texttt{accidentalSharpBad} and \texttt{gClefBad}). This adjustment causes the classifier loss to be affected when a deformed symbol is generated, as these images are now classified under the "bad" classes rather than their correct counterparts.

A second issue is evident in Fig. ~\ref{fig:hw1}.b, where, despite significant variation in the input images of the quarter notes, the generated synthetic images exhibit a high degree of similarity. This behavior suggests that the GAN is excessively dependent on the input text label, leading to nearly equivalent outputs for each instance of a given class while neglecting the style reference image. To address this, we introduce a random swap of a subset of the input labels within a batch. This strategy reduces the GAN’s reliance on the input label, as it may occasionally be inaccurate, and encourages the model to focus more on the input image. As a result, this modification enhances the variability of generated instances within the same symbol class.

As the GAN continues its training process, determining the best time to freeze its weights and save the synthetic symbols to put them into musical staves is required. A first approach to automate this process consists on comparing the encoded feature vectors of the input and generated images using Euclidean Distance and Cosine Similarity. 
%Euclidean Distance measures the absolute difference between the vectors, while Cosine Similarity focuses on the angle between them, emphasizing directional similarity over magnitude.
However, using this decision criterion alone, the generated images do not conform to desired quality standards. To solve this, the Structural Similarity Index Metric (SSIM) is additionally incorporated, which considers structural information, luminance, and contrast to better align with human visual perception. The combination of these metrics allows saving models that generate higher-quality images.

\subsection{Line-level Generation} \label{subsec:sheetgeneration}

Once the synthetic music symbol images are generated, they need to be placed onto music sheets for testing purposes. This task is accomplished using the Smashcima software \cite{mayer2021synthesizing}, which operates as follows: first, the user provides a MusicXML file, which describes the semantics of the music sheet to engrave. Smashcima then collects musical symbol samples from the MUSCIMA++ dataset, as well as their masks, bounding boxes and other positional parameters. The software then places the symbols on the score according to the MusicXML file, resizing and merging them accordingly. For this work, the Smashcima is modified to take the symbol-level images generated by our system rather than the data from MUSCIMA++.

\subsection{Considered Metrics} \label{subsec:consideredmetrics}

For the evaluation of the synthetic musical lines containing our symbols, we utilize the methodology proposed by Pippi et al. \cite{pippi2023hwd}, which provides a tool for comparing two image datasets—music sheets at the line level in our case—by computing multiple metrics in their latent spaces. The following metrics are selected to quantitatively assess the realism of the generated data. Note that in all of these metrics, the lower the value the better.

\tit{Fréchet Inception Distance (FID) \cite{heusel2017gans}} It measures the similarity between two datasets by extracting feature embeddings from a pretrained InceptionV3 model and computing the Fréchet distance between their distributions, assuming a Gaussian form.
\tit{Kernel Inception Distance (KID) \cite{binkowski2018demystifying}} Extracts feature embeddings using InceptionV3, like FID, but measures similarity with Maximum Mean Discrepancy (MMD) using a polynomial kernel. Unlike FID, KID does not assume a Gaussian distribution, making it more robust to different data distributions within the data.
\tit{Handwriting Distance (HWD) \cite{pippi2023hwd}} A metric for Handwritten Text Generation (HTG) that uses a deep network to extract handwriting style features and compute perceptual distance between styles.

To evaluate these metrics, we conducted a comparative analysis of our synthetic lines against multiple reference datasets. Specifically, we compared them to lines generated using the base Smashcima software with Muscima++ symbols, real historical lines, and printed lines, all benchmarked against the Muscima++ dataset.

\begin{figure}[htbp]
    \centerline{\includegraphics[width=0.4\columnwidth]{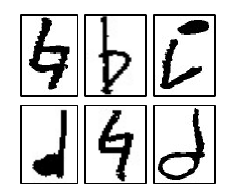}}
    \caption{Mix of real and synthetic symbols. Check footnote $^3$ to know which ones are generated.}
    \label{fig:results_symbols_good}
\end{figure}

\section{Evaluation}

In this section, we evaluate our GAN-generated symbols and staves, both qualitatively and quantitatively. 

\subsection{Qualitative Results}

For the qualitative analysis, we first evaluate the individual handwritten symbols, followed by an assessment of the lines engraved using these synthetic symbols.

As illustrated in Fig. ~\ref{fig:hw1}.a, the majority of the generated symbols are realistic, recognizable, and visually appealing. To further demonstrate this, a mix of real and synthetic symbols is shown in Fig. ~\ref{fig:results_symbols_good}. The generated symbols are revealed in the footnote \footnote{Images generated by the GAN: top-left, top-right, bottom-right.}.

\begin{figure}[htbp]
    \centerline{\includegraphics[width=0.4\columnwidth]{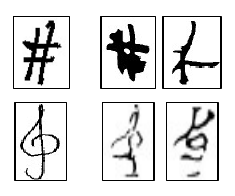}}
    \caption{On the left, reference images of an \texttt{accidentalSharp} and a \texttt{gClef} from the input dataset. On the right, faulty images generated by the GAN of each respective class.}
    \label{fig:results_symbols_bad}
\end{figure}

However, certain symbols remain challenging for the GAN to reproduce accurately, particularly the \texttt{accidentalSharp} and \texttt{gClef} classes. Nevertheless, as shown in Fig. ~\ref{fig:results_symbols_bad}, while not of perfect quality, the generated instances start to ressemble more the real symbols.

\begin{figure}[!ht]
    \centering
    \includegraphics[height=1.5cm]{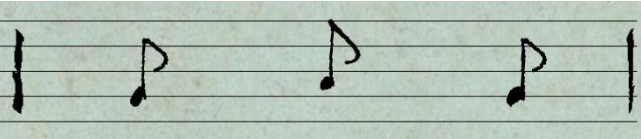} \\[0.1cm]
    \includegraphics[height=1.5cm]{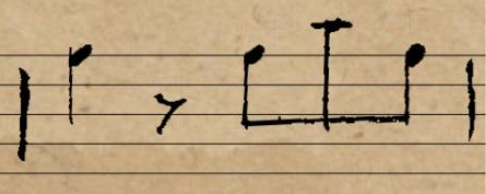}\\[0.1cm]
    \includegraphics[height=1.5cm]{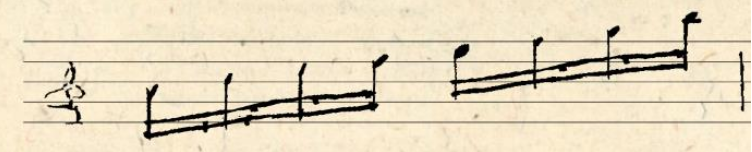}
    \caption{Musical sheets made with the symbols generated by the GAN, then
cut into a measure and cropped vertically.}
    \label{fig:smashcima1}
\end{figure}

Regarding the staves generated using our synthetic symbols, some examples are shown in  Fig. ~\ref{fig:smashcima1}. Overall,the backgrounds utilized by Smashcima effectively replicate various types of aged paper, while the connection between primitives is well-executed.

Some comparison can be drawn between our approach and that of Tirupati \etal \cite{tirupati2024crafting}, since both methods employ GANs in different ways for music score generation. One of the key strengths of their work is its ability to closely replicate the musical structure from the input images. However, this also introduces a limitation, as the generated scores closely adhere to the style of the original input music sheet, restricting stylistic variability and the model's freedom to generate interesting handwriting artifacts. We believe that an ideal synthetic music score generation system should allow for the creation of any music sheet, in the style of any specific composer while maintaining a high degree of realism. Our proposed method, which first generates the primitives and then arranges them on the page, has the advantage of letting us control the musical structure without being limited by the input data. %Nevertheless, it lacks the style consistency achieved in Tirupati's work. Luckily, this may be achieved by either upgrading the GAN architecture so it generates images precisely replicating styles of specific authors, or applying a postprocessing step where the generated images are clustered by similarity of handwritten styles. This would allow us to craft any music sheet from which we have a MusicXML file, with any handwritten style learned by the GAN, and, in some cases, realistic enough to seem as if written by a human.

\subsection{Quantitative Results}

Following a visual assessment of the generated musical lines, their accuracy was quantitatively evaluated by comparing them to real lines extracted from the MUSCIMA++ dataset. In addition to our proposed dataset -- referred to as \texttt{GAN Smashcima} -- several other line-level datasets were included for comparison. These comprise \texttt{Base Smashcima}, which contains lines generated using the original Smashcima software based on MUSCIMA++ symbol masks; \texttt{Real}, which includes authentic historical manuscript lines; and \texttt{Printed}, consisting of typeset musical lines derived from printed scores. All datasets were binarized prior to comparison to ensure that evaluations focused exclusively on the stroke and structural form of the symbols, rather than on background similarity.

\begin{comment}
\begin{figure}[!ht]
    \centerline{\includesvg[width=1.10\columnwidth]{Materials_paper/muscima_fid.svg}}
    \centerline{\includesvg[width=1.10\columnwidth]{Materials_paper/muscima_kid.svg}}
    \centerline{\includesvg[width=1.10\columnwidth]{Materials_paper/muscima_hwd.svg}}
    \caption{Fréchet Inception Distance (FID), Kernel Inception Distance (KID), and Handwriting Distance (HWD) between various datasets and the MUSCIMA++ dataset as reference, which consists of real handwritten lines.}
    \label{fig:charts}
\end{figure}
\end{comment}

\begin{table*}[!ht]
    \centering

    \renewcommand{\arraystretch}{1.5}
    \begin{tabular}{@{}l@{\hspace{20pt}}c@{\hspace{20pt}}c@{\hspace{20pt}}c@{}}
        & \textbf{FID} $\downarrow$ & \textbf{KID} $\downarrow$ & \textbf{HWD} $\downarrow$ \\
        \toprule
        \textbf{Printed} & 118.96 & 0.114 & 2.194 \\
        \textbf{Real} & 136.82 & 0.117 & 2.32 \\
        \textbf{Base Smashcima} & 132.38 & 0.128 & 3.052 \\
        \textbf{GAN Smashcima} & 142.48 & 0.150 & 2.85 \\
        \midrule
        \textbf{MUSCIMA++ (Self)} & \textbf{0.033} & \textbf{-1.06} & \textbf{0.0} \\
        \bottomrule
    \end{tabular}
    \vspace{1.75em}
    \caption{Fréchet Inception Distance (FID), Kernel Inception Distance (KID), and Handwriting Distance (HWD) between various datasets and the MUSCIMA++ dataset as reference, which consists of real handwritten lines.}
    \label{tab:dataset_comparison}
\end{table*}

Table ~\ref{tab:dataset_comparison} reveals that the dataset most similar to MUSCIMA++ across all metrics is the printed dataset. %, for which a higher (worse) score was expected, therefore, this suggests an avenue for further research. 
According to the FID metric, Base Smashcima follows, which is consistent given that its symbols originate from MUSCIMA++. The real dataset ranks next, as expected, since it consists of human-written symbols. Finally, the dataset generated in this work achieves an FID score close to the real dataset, which indicates that the global structure of our images closely resembles MUSCIMA++.

The KID metric, also in Table ~\ref{tab:dataset_comparison}, exhibits a similar ranking, with the only difference being that real lines outperform those from Base Smashcima. This may be attributed to the greater stability of the KID metric with respect to the number of samples, given that the number of lines generated for this test is approximately in the thousands.

For the HWD metric, illustrated in the last chart in Table ~\ref{tab:dataset_comparison}, the printed and real datasets again yield the best results. Notably, our generated symbols score higher in handwriting similarity than Base Smashcima, despite the latter being derived from MUSCIMA++. This suggests that, in the absence of real data, our synthetic symbols may serve as a viable alternative for augmenting datasets.

To better contextualize the significance of the differences observed in the evaluation metrics, we established a baseline by comparing the Muscima++ dataset against itself. As expected, the HWD metric yielded a score of 0.0, indicating perfect similarity. The FID metric produced a low score of 0.033, while the KID metric resulted in a slightly negative value of -1.06. Although the KID metric is theoretically non-negative—as it estimates a squared distance—it is an unbiased estimator and can exhibit slight negative values in practice due to variance introduced by finite sample sizes.

\section{Conclusions}

This work has proposed a GAN architecture for generating synthetic handwritten musical symbols. We have demonstrated that, even in a field characterized by limited and complex structured data, a GAN-based model can produce promising results, offering a new avenue for providing training data for OMR. Furthermore, we have explored the integration of the Smashcima software for engraving synthetic symbols, which proved effective in generating realistic music scores while allowing precise control over the desired musical content.

Our pipeline opens several possibilities for future research. First, by leveraging our GAN-generated symbols, researchers can explore other techniques for arranging them into realistic sheets. Second, different generative architectures could be tested, such as Diffusion Models or Transformers, which have not been explored in the field of OMR yet, and then integrated into our existing pipeline, and using the final processing stage to structure the generated symbols into coherent musical notation.

Despite the promising results, opportunities for further improvement remain. The generation of some symbols (i.e. clef and accidental sharp) is challenging due to their inherent complexity and the high handwriting style variability. This could be addressed by making the GAN learn the handwritten style of a specific author, whenever data of distinct writers may be available. Another limitation is the lack of symbol classes that the Smashcima software accepts and includes in the generated scores. If the number of classes accepted were to be increased, more complex music sheets could be generated, and therefore, be able to get closer to the feature representation of the real data.

%ORIGINAL: Despite the promising results, opportunities for further improvement remain. The generation of the two most challenging symbols, as discussed throughout this work, still exhibits some inconsistencies. Our current hypothesis attributes this challenge to the inherent complexity of these symbols and the resulting variability in their handwritten representations. This could be addressed by making the GAN learn the handwritten style of a specific author. We leave this for future work where more data of distinct authors may be available. Another limitation is the lack of symbol classes that the Smashcima software accepts and includes in the generated scores. If the number of classes accepted were to be increased, more complex music sheets could be generated, and therefore, be able to get closer to the feature representation of the real data.

Addressing these challenges could bring us closer to high-quality synthetic handwritten notation. As generative models and domain adaptation advance, synthetic data is likely to play a crucial role in bridging the gap between handwritten and machine-readable notation.

\begin{credits}
\subsubsection{\ackname} 

This work has been partially supported by the Spanish projects
CNS2022-135947 (DOLORES) and PID2021-126808OB-I00 (GRAIL) from the Minis-
terio de Ciencia e Innovación, and the grant Càtedra ENIA UAB-Cruïlla (TSI-100929-2023-
2) from the Ministry of Economic Affairs and Digital Transition of Spain. Pau Torras is funded by the Spanish FPU Grant FPU22/00207.

\subsubsection{\discintname}

The authors have no competing interests to declare that are relevant to the content of this article.

\subsubsection{Code Availability}

\leavevmode\\\noindent
Link to the GAN code: \url{https://github.com/GerardAsbert/CVC_OMR-GAN}.\\
Link to the custom Smashcima code: \url{https://github.com/GerardAsbert/Smashcima}

\end{credits}
\bibliographystyle{splncs04}
\bibliography{paper_GAN}

\end{document}